\documentclass[runningheads]{llncs}

\usepackage[T1]{fontenc}
\usepackage{xcolor}
\usepackage{graphicx}
\usepackage{hyperref}
\usepackage{amsmath}
\usepackage{amssymb}
\usepackage{cleveref}
\usepackage{multicol}
\usepackage{multirow}
\usepackage{booktabs}

\begin{document}


\title{RS-OVC: Open-Vocabulary Counting for Remote-Sensing Data}

\author{
Tamir Shor\inst{1,2}\thanks{This work was done during an internship at Google Research}\orcidID{0009-0008-9537-2558} \and
George Leifman\inst{2}\orcidID{0000-0003-0819-1692} \and
Genady Beryozkin\inst{2}\orcidID{0009-0003-4722-7640}
}

\authorrunning{T. Shor et al.}

\institute{
Technion -- Israel Institute of Technology \and
Google Research \\
\email{tamir.shor@campus.technion.ac.il} \\
\email{\{gleifman,genady\}@google.com}
}

\hypersetup{
    colorlinks=true,
    urlcolor=purple
}

\maketitle

\begingroup
\renewcommand\thefootnote{**}
\footnotetext{Code is available at \href{https://github.com/tamirshor7/RS-OVC}{\textcolor{purple}{https://github.com/tamirshor7/RS-OVC}}}
\endgroup

%

\begin{figure*}
    \centering
    \includegraphics[width=0.9\linewidth]{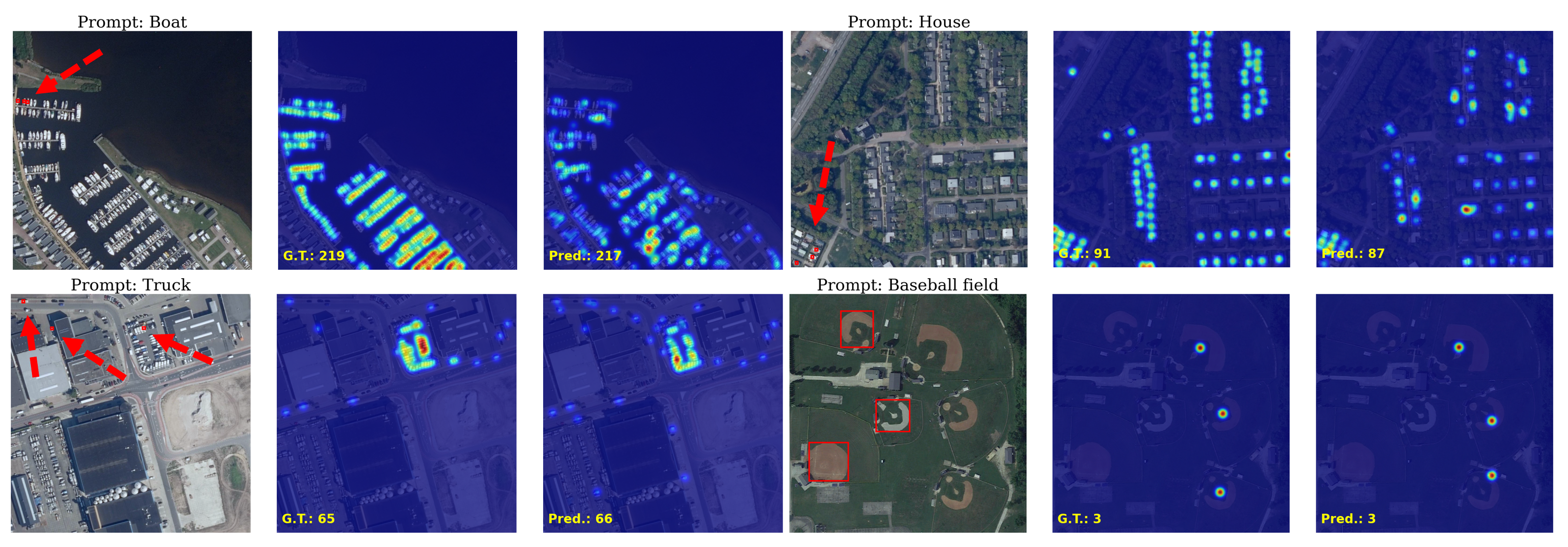}
    \caption{\textbf{Object confidence maps} - illustrating spatial correspondence with aggregated textual \textit{(i.e. prompt)} and visual \textit{(i.e. exemplar)} conditioning. Visual exemplars are marked in red bounding-boxes on each image (red arrows highlight exemplars).}
    \label{fig:teaser}
\end{figure*}

\begin{abstract}
Object-Counting for remote-sensing (RS) imagery is attracting increasing research interest due to its crucial role in a wide and diverse set of applications. While several promising methods for RS object-counting have been proposed, existing methods focus on a closed, pre-defined set of object classes. This limitation necessitates costly re-annotation and model re-training to adapt current approaches for counting of novel objects that have not been seen during training, and severely inhibits their application in dynamic, real-world monitoring scenarios.
To address this gap, in this work we propose RS-OVC - the first Open Vocabulary Counting (OVC) model for Remote-Sensing and aerial imagery. We show that our model is capable of accurate counting of novel object classes, that were unseen during training, based solely on textual and/or visual conditioning.
\end{abstract}


\section{Introduction}
\label{intro}
Object counting stands at the core of numerous remote-sensing (RS) applications, providing essential quantitative information for understanding, managing, and responding to large-scale real-world phenomena. RS counting has proven useful in disaster control \cite{fan2017quantifying,barrington2011crowdsourcing}, urban planning \cite{garrido2023people,michelat2010automatic,angeoletto2018counting}, wildlife-tracking \cite{hollings2018you,arteta2016counting}, crowd-monitoring \cite{song2012real,saleh2015recent,guo2022density}, maritime surveillance \cite{reggiannini2024remote,park2023aerial}, ecological conservation \cite{bernd2017more,senf2022seeing}, and in a myriad of other tasks \cite{dare2002application,farjon2023deep,chong2017review}. 
The demand for accurate and automated counting methods is on the rise due to increasing challenges in the aforementioned (and other) applications, as well as due to growth in potential volume and resolution of available RS data.\\
These trends have fostered a growing amount of research dedicated towards the specific challenges that arise in object-counting for aerial imagery. Prior work shows encouraging developments tackling challenges of low-resolution and noisy background that often appear in aerial imagery \cite{gao2020counting,guo2022density,guo2024balanced}, as well as considerations related to resource consumption for scalable deployment on edge-devices \cite{duan2021distillation,shen2024lightweight,li2023instance}.
The main limitation of current related research we address in this paper is that existing approaches for RS object-counting are heavily-focused on \textit{closed-set} object counting. Namely, current methods are trained to count object instances from a finite set of classes that is pre-defined during training. These methods disallow the counting of objects from novel classes, which have not been seen at train time. Adaptation of existing approaches to such new object classes would require costly data annotation and model training. This makes current paradigms fall short in dynamic, real-world scenarios where RS object counting would be most relevant. \\
To bridge this gap, in this work we propose a model for Remote-Sensing Open-Vocabulary Counting (RS-OVC). The development of RS-OVC is predicated on identifying and resolving two key challenges. First, training robust OVC models demands not only large-scale annotation but also diverse object categories, a requirement that existing RS counting datasets typically fail to meet, being largely constrained to a limited set of object classes. Second, Open-Vocabulary Counting (OVC) is intrinsically more difficult than its closed-set counterpart, requiring the model to generalize counting knowledge to unseen classes and to avoid bias towards objects seen at train time.\\
We overcome the first challenge by efficiently leveraging and combining knowledge from pre-trained vision encoders for both general Computer Vision (CV) and Remote Sensing (RS), and by curating and consolidating several diverse RS detection datasets to create the necessary comprehensive training environment for open-vocabulary generalization. The second mentioned challenge is addressed by building upon CountGD \cite{amini2024countgd}- an OVC model from general computer-vision. While designing custom architectures is common in Remote-Sensing literature for object counting \cite{gao2020counting,guo2022object,liu2025mamba}, in this work we take a different path. Our motivation for building upon the existing CountGD framework, and general rationale in this work, is simplicity. Namely, this work is guided by the principle of creating an accurate OVC model for Remote-Sensing data with minimal adaptations to existing viable solutions.

\section{Related Work}
\subsection{Object-Counting in Remote-Sensing}
Object-counting is trivially attained with object-detection models. The main motivation for the development of algorithms dedicated specifically for object counting is that object detection models struggle in scenes that exhibit high scale variation, high object density and low-resolution imagery. 
The focus on counting enables smoothed modeling of the output space (e.g. via density estimation \cite{ding2022object,klemela2009smoothing}) and optimization that allows increased focus dedicated to the aforementioned challenges faced by detection models. These principles have guided several prominent efforts in RS and general computer-vision object counting in recent years.\\
MCNN \cite{zhang2016single} proposed density estimation based on multi-scale convolutions to cope with scale variations. ASPDNet \cite{gao2020counting} is one of the first works tackling object-counting specifically for RS data. This method copes with scale and orientation variation by utilizing a pyramid feature network (PFN), and background noise is addressed with a designated spatial-attention for importance-weighting in input space. TASNet \cite{guo2022object} employs a similar approach integrated with multi-channel attention to suppress background clutter. EdgeCount\cite{shen2024lightweight} exhibits an additional advantage of object counting for RS, demonstrating accurate counting can be obtained using highly-compact models via information distillation. Recent works have also successfully leveraged Vision-Transformers (ViTs)\cite{shen2024lightweight,chen2022transformer} and State-Space models \cite{liu2025mamba} for RS-object counting. As previously mentioned - all methods surveyed here are focused on closed-set object counting.

\subsection{Open-Vocabulary Counting}
While the field of Open-Vocabulary Detection (OVD) has seen rapid progress in both CV\cite{liu2024grounding} and RS \cite{wang2025mask,pan2025locate}, research dedicated specifically to Open-Vocabulary Counting (OVC) remains limited in both fields of research. Leveraging OVD models for counting is possible, however this method is suboptimal due to previously-mentioned difficulties of detection models in highly dense scenes.

To develop an OVC solution for RS data, in this work we adapt \textit{CountGD} \cite{amini2024countgd}, an OVC solution originally developed for general computer-vision. CountGD achieves open-set counting by reformulating the task as a grounding problem built upon the DINO architecture \cite{liu2024grounding}. CountGD interleaves language features from a BERT text-encoder \cite{devlin2019bert} and (optionally) visual-exemplar features from a Swin Transformer \cite{liu2021swin} using self-attention. These features are cross-attended with the $900$ image features that exhibit highest similarity to the fused conditioning features to create the final output similarity scores used for counting.

\section{Method}
The key challenge guiding our work is that OVC requires large-scale diverse data for training. While there are several publicly-available annotated datasets for RS counting (e.g. RSOC \cite{gao2020counting}, VHR-Ships \cite{kizilkaya2022vhrships}, NWPU-MOC \cite{gao2024nwpu}), these datasets lack the category diversity, that is often available when training OVC and OVD models for general CV (e.g. COCO \cite{lin2014microsoft}, PASCAL-VOC \cite{everingham2010pascal}). To cope with these challenges, our approach in this work is to utilize learned fusion of textual and visual features acquired in pre-training an OVC model over rich general CV data, and adapt this knowledge onto the RS domain.\\
The following section lays out our method - in \Cref{countgd} we survey the CountGD backbone utilized in our model. Section \ref{rfi} presents our approach for injecting RS features into the pre-trained countGD components. Section \ref{data} elaborates on our training data and contrastive optimization regime for promoting zero-shot and few-shot counting capabilities.
\subsection{CountGD Backbone}
\label{countgd}

CountGD~\cite{amini2024countgd} is a multi-modal framework for open-world object counting, which allows specifying the target object through either text prompts, visual exemplars (visual examples of the object of interest, supplied by the user as bounding boxes on the input image), or both jointly. 
The model extends the open-vocabulary detector GroundingDINO~\cite{liu2024grounding} by introducing modules that encode visual exemplars alongside text. The architecture includes: (i) an image encoder $f_{\theta}^{\text{SwinT}}$, (ii) a text encoder $f_{\theta}^{T}$, (iii) a feature enhancer $f_{\phi}$, (iv) a language- and exemplar-guided query selection mechanism, and (v) a cross-modality decoder $f_{\psi}$.\\

The image encoder $f_{\theta}^{\text{SwinT}}$ (a Swin Transformer~\cite{liu2021swin}) extracts multi-scale feature maps from a given input image $X\in\mathbb{R}^{H \times W \times 3}$ , projected into image tokens $z_I$. Visual exemplar tokens $z_v$ are obtained by applying RoIAlign~\cite{ren2016faster} to these feature maps at the exemplar coordinates $B$. The text encoder $f_{\theta}^{T}$ (BERT\cite{devlin2019bert}-based) encodes the textual description $t$ into textual features $z_t$.

For $n$ image tokens, $p$ visual exemplar tokens, and $q$ text tokens, the feature enhancer $f_{\phi}$ fuses the image, text, and exemplar features through self- and cross-attention. It outputs fused exemplar–text features $z_{v,t}$ and enhanced image features $z_I$ as
\[
(z_{v,t}, z_I) = f_{\phi}\big(f_{\theta}^{\text{SwinT}}(X), \text{RoIAlign}(f_{\theta}^{\text{SwinT}}(X), B), f_{\theta}^{T}(t)\big).
\]

The \textit{Select} operation elects the top-$k=900$ image tokens most similar to the fused features, and those features are fed as queries to the cross-modality decoder $f_{\psi}$, which computes the similarity matrix:
\[
\hat{Y} = \sigma\big(f_{\psi}(z_I, z_{v,t}, \text{Select}(z_I, z_I z_{v,t}^{\top}, k)) \, z_{v,t}^{\top}\big) \in \mathbb{R}^{k\times(p+q)},
\]
where $z_I z_{v,t}^{\top} \in \mathbb{R}^{n \times (p+q)}$ represents the similarity between $n$ image tokens and $p+q$ text/exemplar tokens. $\sigma(\cdot)$ denotes the element-wise sigmoid, maintaining confidence scores that indicate the likelihood of a candidate image feature corresponding to a given prompt token. Tokens exceeding a confidence threshold are enumerated to yield the final count $\hat{y}$.
During training, the text encoder $f_\theta^{T}$ and image encoder $f_\theta^{\mathrm{SwinT}}$ remain frozen, while the exemplar projection layer $f_\varphi$, feature enhancer $f_\phi$, and decoder $f_\psi$ are optimized according to a localization and classification objective:
\[
\mathcal{L} = \lambda_{\text{loc}} \mathcal{L}_{\text{loc}} + \lambda_{\text{cls}} \mathcal{L}_{\text{cls}},
\]
where $\mathcal{L}_{\text{loc}}$ is an L1 loss on predicted box centers and $\mathcal{L}_{\text{cls}}$ is a focal loss on the similarity matrix $\hat{Y}$. 

\subsection{RS Feature Injection}
\label{rfi}
As further demonstrated in \Cref{results}, adapting CountGD for Remote-Sensing data by straightforward finetuning over RS data leads to suboptimal downstream performance. We claim the reason for this is that existing datasets for RS counting are characterized by lack of object diversity in highly dense aerial images, that ultimately leads to mode-collapse and catastrophic forgetting at train time. \\
To avoid discarding rich feature representation and knowledge gathered for feature aggregation during the training of countGD over general CV data, in this work we propose an alternative approach - in RS-OVC we perform ad-hoc injection of RS features into the feature space of the existing pre-trained image and text encoders of the native CountGD architecture. To this end we utilize the DINOv3 encoder\cite{simeoni2025dinov3} pre-trained on RS data. Each input image $X$ (including the image exemplars) is encoded by both encoders independently to receive Swin features (pre-trained on diverse CV data) $z_{CV}=f_{\theta_1}^{\text{SwinT}}(X)$ and ViT features (pre-trained on RS data) $z_{RS}=f_{\theta_2}^{\text{DINO}}(X)$. We cross-attend every $z_{CV}$ (used as keys and values) with $z_{RS}$ (as queries) for output features from every resolution level of the Swin Transformer encoder, receiving a fused feature representation for every resolution level of $f_{\theta_1}^{\text{SwinT}}$:
\[
z_{fused} = CrossAttention(K,V = z_{CV} ; Q = z_{RS}). 
\]
This RS-informed set of image features is used for the feature fusion and query selection modules used in CountGD:
\[
(z_{v,t}, z_I) = f_{\phi}\big(z_{fused}, \text{RoIAlign}(z_{fused}, B), f_{\theta}^{T}(t)\big).
\]
Importantly, at train time we freeze both image encoders and the text encoders. This allows useful preservation of pre-training knowledge, and also substantially reduces computational resource consumption and optimization instability during training.  Our pipeline is visually-presented in \Cref{fig:model}. Additional implementation and optimization details are reported in the supplementary.
\begin{figure}
    \centering
    \includegraphics[width=1\linewidth]{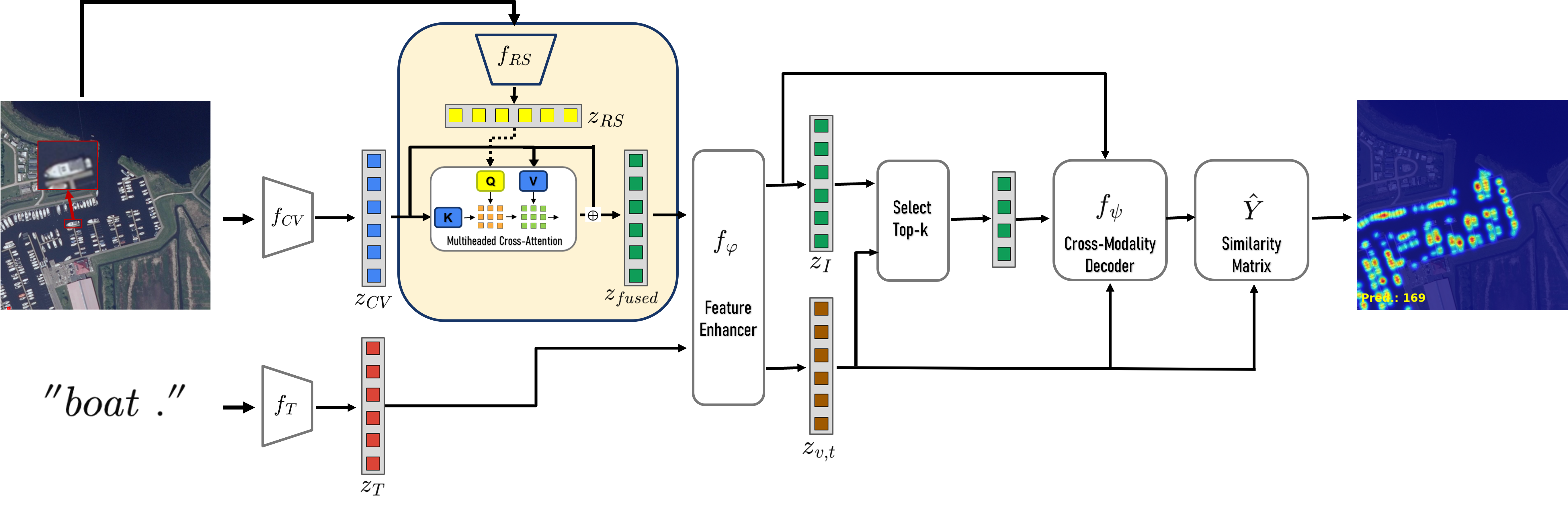}
    \caption{\textbf{RS-OVC Pipeline} - Our modifications from the original CountGD architecture are highlighted with an orange background. Image and text encoders remain frozen during optimization, other parameters are finetuned.}
    \label{fig:model}
\end{figure}
\subsection{Data Curation \& Training}
\label{data}
To address the challenge of insufficient object and scale diversity for training an open-world object-counting model, we enrich existing RS counting datasets with large-scale, category-diverse RS detection datasets. Specifically, we adapt FAIR1M~\cite{sun2022fair1m}, DIOR~\cite{li2020object}, and DOTA~\cite{xia2018dota} to the OVC setting by converting all annotated bounding boxes into point-based instance labels using their centroid coordinates. These datasets are combined with NWPU-MOC~\cite{gao2024nwpu}, a diverse RS counting benchmark containing 14 object categories.

\begin{figure}
    \centering
    \includegraphics[width=1\linewidth]{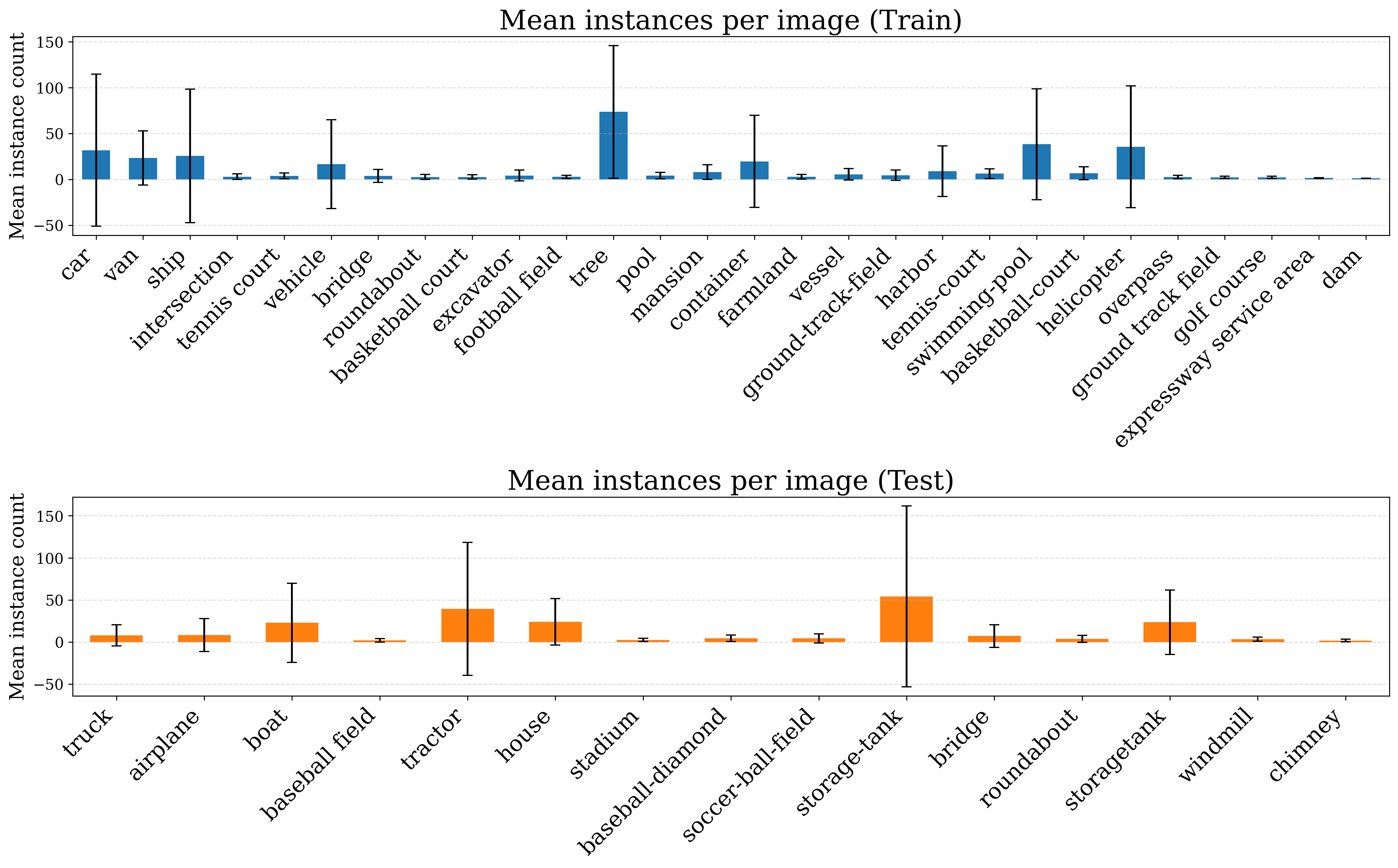}
    \caption{Curated dataset class-wise mean and standard-deviation (error bars) for object instance counts across images, for the training (top) and test (bottom) splits.}
    \label{fig:data}
\end{figure}

For each dataset, we decompose multi-class images into separate samples, ensuring that each training instance contains only a single object class. From this pool, we retain only samples containing at least four instances of the target class, enabling 3-shot conditioning during training -- 3 annotations are used for visual exemplars, remaining annotations are used for ground-truth object counts. Finally, we construct a unified train–validation–test split over the merged set of classes from all curated datasets. The full composition of object classes and object density of our curated dataset is fully-shown in \Cref{fig:data}. Importantly, to measure few-shot OVC performance, all object classes used for testing have not been used as any form of conditioning (textual, visual or other) during training or validation. The \textit{harbor} and \textit{helicopter} classes are excluded for validation of architecture design choices. Additionally, for each dataset we exclude  100 random samples from the training set to determine the counting confidence threshold. 

\section{Results}
\label{results}
\subsection{Quantitative Results}
\label{quant}
In this section we report Mean Absolute Error (MAE) and Root Mean Squared Error (RMSE) of RS-OVC's counting errors across held-out test classes from our curated NWPU-MOC\cite{gao2024nwpu} (\Cref{tab:nwpu_results}), FAIR-1M\cite{sun2022fair1m} (\Cref{tab:fair_results}), DOTA\cite{xia2018dota} (\Cref{tab:dota_results}), DIOR\cite{li2020object} (\Cref{tab:dior_results}) datasets. In \cref{tab:rsocb_results} we also evaluate our method over the RSOC\cite{gao2020counting} dataset's building class.\\
Given that we are the first to propose an OVC model for RS data, we measure our model's performance against an existing RS open-vocabulary detection (OVD) model, and against an OVC model developed for general CV data. As an OVD baseline we compare to Locate-Anything-on-Earth (LAE) \cite{pan2025locate}, which is a SOTA RS open-vocabulary object-detection model. This model can be trivially adapted for object counting by using the number of detected bounding-boxes per-prompt. For an OVC baseline we compare against the original CountGD \cite{amini2024countgd} model. For CountGD, we evaluate both the off-the-shelf model pre-trained on CV data, applied directly to RS data (denoted "RS-FT" in the tables). Furthermore, to illustrate the advantage of the finetuning approach proposed in our work, we compare to a version of CountGD finetuned on our curated RS dataset, without the additional RS feature fusion of RS-OVC. Lastly, to demonstrate the issue of catastrophic forgetting and importance of our finetuning and feature fusion, we ablate between the two options of (i) using a frozen RS encoder alongside a pre-trained CV Swin transformer-encoder (\Cref{rfi}) and (ii) the alternative of using only a frozen RS image encoder (denoted \textit{"RS-only"}), relinquishing CountGD's pre-trained image encoder and any feature fusion logic.

\begin{table}[t]
\centering
\setlength{\tabcolsep}{2pt} 
\renewcommand{\arraystretch}{1.2}
\caption{Object-Counting performance on NWPU-MOC test classes. \textit{CountGD} indicates frozen off-the-shelf pretrained CountGD , \textit{RS-FT} indicates original pretrained CountGD finetuned over RS data, \textit{RS-only} indicates RS-OVC using only an RS pretrained image encoder, without feature fusion. }
\resizebox{\linewidth}{!}{%
    
    \begin{tabular}{|l|cc|cc|cc|cc|cc|cc|} 
    \hline
    \textbf{Method}
    & \multicolumn{2}{c|}{Boat}
    & \multicolumn{2}{c|}{House}
    & \multicolumn{2}{c|}{Truck}
    & \multicolumn{2}{c|}{Stadium}
    & \multicolumn{2}{c|}{Airplane}
    & \multicolumn{2}{c|}{Mean} \\
    \cline{2-13}
    & MAE & RMSE & MAE & RMSE & MAE & RMSE & MAE & RMSE & MAE & RMSE & MAE & RMSE \\
    \hline
    CountGD 
    & 32.58 & 63.56 & 23.54 & 35.66 & 11.38 & 29.59 & 2.5 & 3.16 & 1.97 & 2.52 & 20.16 & 37.1\\
    \hline
    CountGD (RS-FT)
    & 36.25 & 76.01 & 21.75 & 34.88 & 8.80 & 18.58 & \textbf{1.45} & \textbf{2.13} & \textbf{1.19} & \textbf{1.85} & 18.71 & 36.89 \\
    \hline
    LAE 
    & 40.03 & 77.90 & 27.11 & 38.73 & 12.72 & 20.70 & 5.50 & 5.83 & 4.96 & 5.21 & 23.44 & 39.73 \\
    \hline
    RS-OVC (RS-only)
    & 32.47 & 72.56 & 18.56 & 31.13 & 10.21 & \textbf{17.32} & 4 & 4.69 & 6.5 & 7.0 & 17.07 & 34.01\\
    \hline
    RS-OVC (Ours)
    & \textbf{7.5} & \textbf{15.44} & \textbf{15.52} & \textbf{26.87} & \textbf{8.74} & 24.2 & 5.52 & 6.37 & 3.08 & 3.55 & \textbf{12.42} & \textbf{24.72}\\
    \hline
    \end{tabular}%
}

\label{tab:nwpu_results}
\end{table}

\begin{table}[t]
\centering
\footnotesize
\setlength{\tabcolsep}{4pt}
\renewcommand{\arraystretch}{1.2}
\caption{Object-Counting performance on FAIR-1M test classes.}
\resizebox{\linewidth}{!}{

\begin{tabular}{|l|cc|cc|cc|cc|cc|cc|cc|}
\toprule
\textbf{Method}
& \multicolumn{2}{c|}{Tractor}
& \multicolumn{2}{c|}{Boat}
& \multicolumn{2}{c|}{Truck}
& \multicolumn{2}{c|}{Airplane}
& \multicolumn{2}{c|}{Baseball-Field}
& \multicolumn{2}{c|}{Mean} \\
\cline{2-13}
& MAE & RMSE & MAE & RMSE & MAE & RMSE & MAE & RMSE & MAE & RMSE & MAE & RMSE  \\
\midrule
CountGD 
& 52.74 & 73.42 & 18.22 & 35.57 & 39.78 & 55.90 & 3.89 & 9.93 & 7.61 & 14.81 & 21.65 & 39.93 \\
\hline
CountGD (RS-FT)
& 37.03 & 86.25 & 18.12 & 45.42 & 5.79 & 12.61 & 4.27 & 9.64 & 1.39 & 1.68 & 6.57 & 19.06 \\
\hline
LAE
& 41.58 & 89.57 & 19.16 & 45.53 & 10.65 & 15.88 & \textbf{1.29} & \textbf{6.74} & \textbf{0.48} & \textbf{0.82} & 7.56 & 19.68 \\
\hline
RS-OVC (RS-only)
& 37.55 & 86.66 & 18.14 & 45.23 & 6.04 & 12.72 & 4.39 & 9.75 & 1.58 & 1.88 & 6.74 & 19.08 \\
\hline
RS-OVC (Ours)
& \textbf{26.84} & \textbf{47.98} & \textbf{12.79} & \textbf{27.45} & \textbf{5.54} & \textbf{10.41} & 4.21 & 10.22 & 3.00 & 3.06 & \textbf{5.81} & \textbf{13.58} \\

\bottomrule
\end{tabular}
}

\label{tab:fair_results}
\end{table}

Table \ref{tab:nwpu_results} reports the MAE and RMSE over the NWPU-MOC\cite{gao2024nwpu} dataset. \textit{Mean} column indicates mean test results across the entire test set. Results indicate our model outperforms all considered baselines. An exception is the "stadium" and "airplane" classes, where finetuned CountGD produces better results than our model. We believe that this is because the "airplane" and "stadium" classes possess distinct features that mostly appear in larger scales. These features, which are less specific to remote sensing, require less dedicated adaptation of CountGD to this domain. In most cases, especially in dense scenes with high counts of small objects or low-resolution objects (e.g. "boat","house", see \Cref{fig:data}), RS-OVC significantly outperforms CountGD variants and the OVD solution from LAE. These conclusions are also shown qualitatively in figure \ref{fig:teaser}, demonstrating the potential of RS-OVC in accurate counting of objects that had not been introduced at train time, even in highly dense and complex scenes.\\
The conclusions are further-supported by \Cref{tab:fair_results}, reporting results over held-out test classes of the FAIR-1M dataset. Our model outperforms all baselines on object classes that tend to appear more-densely in some scenes (e.g. "tractor", "boat", "truck"). For classes with clearer, large-scale features, where objects are less-likely to appear in low-resolutions or high counts (e.g. "stadium,"airplane"), our model showcases comparable yet inferior performance. LAE \cite{pan2025locate} achieves the best results for these object classes. These findings are consistent with the general view in related literature, according to which dedicated object counting models typically outperform detection-based approaches in dense or low-resolution scenarios, whereas detection methods tend to slightly outperform in some of the simpler, sparse scenes. 
Notably, our model surpasses all baselines in mean performance across both tables. This indicates that the substantial gains achieved in complex scenes effectively offset the marginal performance drop observed in sparser scenes compared to LAE and other benchmarks.

\begin{table*}[t]
\centering
\setlength{\tabcolsep}{2pt} 
\renewcommand{\arraystretch}{1.2}
\caption{Object-Counting performance on DOTA test classes.}
\resizebox{\textwidth}{!}{%
    \begin{tabular}{|l|cc|cc|cc|cc|cc|cc|cc|} 
    \hline
    \textbf{Method}
    & \multicolumn{2}{c|}{Storage Tank}
    & \multicolumn{2}{c|}{Airplane}
    & \multicolumn{2}{c|}{Bridge}
    & \multicolumn{2}{c|}{Soccer Ball Field}
    & \multicolumn{2}{c|}{Roundabout}
    & \multicolumn{2}{c|}{Baseball Diamond}
    & \multicolumn{2}{c|}{Mean} \\
    \cline{2-15}
    & MAE & RMSE & MAE & RMSE & MAE & RMSE & MAE & RMSE & MAE & RMSE & MAE & RMSE & MAE & RMSE \\
    \hline
    CountGD 
    & 57.71 & 112.68 & 33.21 & 55.19 & 14.74 & 25.14 & 56.39 & 92.20 & 21.29 & 43.15 & 52.09 & 87.43 & 36.46 & 71.97 \\
    \hline
    CountGD (RS-FT)
    & 51.47 & 116.75 & 35.60 & 77.28 & 11.54 & 20.41 & 10.57 & 28.00 & \textbf{4.18} & \textbf{5.36} & 8.00 & 17.40 & 28.14 & 72.72 \\
    \hline
    LAE
    & 57.18 & 121.67 & \textbf{18.86} & 59.40 & 14.80 & 23.91 & 7.21 & 9.01 & 6.82 & 8.06 & \textbf{4.67} & 6.69 & 23.59 & 67.94 \\
    \hline
    RS-OVC (RS-only)
    & 56.18 & 111.30 & 32.14 & 73.74 & 16.75 & 21.58 & 18.75 & 26.80 & 9.22 & 9.88 & 10.63 & 11.80 & 23.82 & 60.85 \\
    \hline
    RS-OVC (Ours)
    & \textbf{49.76} & \textbf{111.28} & 24.13 & \textbf{41.66} & \textbf{11.51} & \textbf{18.61} & \textbf{5.50} & \textbf{6.27} & 5.67 & 6.93 & 5.33 & \textbf{6.65} & \textbf{23.16} & \textbf{58.15} \\ 
    \hline
    \end{tabular}%
}
\label{tab:dota_results}
\end{table*}

Table \ref{tab:dota_results} exhibits a trend similar to the one from \Cref{tab:nwpu_results,tab:fair_results} - our method outperforms all baselines in mean performance, and most-notably for object classes appearing densely (e.g. "storage tank"). For classes mostly characterized with distinct features or low object-counts, such as "roundabout" and "baseball field", our model shows comparable performance. In \Cref{tab:dior_results} we report counting performance over test-classes from the DIOR dataset. Unlike previous benchmarks used for evaluating our method, DIOR is mostly characterized by relatively sparse scenes with much-lower objects counts. As previously discussed, in such scenes object-detection models tend to outperform models developed strictly for counting. In accordance, \Cref{tab:dior_results} poses DIOR as the only benchmark where our method (and all counting-based baselines) underperforms LAE, which is a detection-based solution.
\begin{table*}[t]
\centering
\setlength{\tabcolsep}{2pt} 
\renewcommand{\arraystretch}{1.2}
\caption{Object-Counting performance on DIOR test classes.}
\resizebox{\textwidth}{!}{%
    \begin{tabular}{|l|cc|cc|cc|cc|cc|cc|cc|cc|} 
    \hline
    \textbf{Method}
    & \multicolumn{2}{c|}{Storage Tank}
    & \multicolumn{2}{c|}{Bridge}
    & \multicolumn{2}{c|}{Airplane}
    & \multicolumn{2}{c|}{Stadium}
    & \multicolumn{2}{c|}{Chimney}
    & \multicolumn{2}{c|}{Baseball Field}
    & \multicolumn{2}{c|}{Windmill}
    & \multicolumn{2}{c|}{Mean} \\
    \cline{2-17}
    & MAE & RMSE & MAE & RMSE & MAE & RMSE & MAE & RMSE & MAE & RMSE & MAE & RMSE & MAE & RMSE & MAE & RMSE \\
    \hline
    CountGD 
    & 14.28 & \textbf{30.17} & 13.93 & 35.29 & 3.43 & 8.67 & 8.00 & 8.00 & 2.02 & 3.61 & 28.00 & 81.35 & 3.10 & 4.08 & 11.70 & 39.02 \\
    \hline
    CountGD (RS-FT)
    & 18.94 & 39.14 & 3.29 & 3.70 & 6.24 & 10.69 & 6.00 & 6.00 & 5.40 & 5.74 & 4.38 & 4.83 & 2.63 & 3.23 & 9.18 & 22.96 \\
    \hline
    LAE
    & \textbf{12.80} & 36.24 & 4.15 & 5.00 & \textbf{1.16} & \textbf{3.24} & \textbf{3.00} & \textbf{3.00} & \textbf{1.09} & \textbf{2.17} & \textbf{1.14} & \textbf{2.79} & \textbf{0.89} & \textbf{2.12} & \textbf{5.05} & \textbf{20.68} \\
    \hline
    RS-OVC (RS-only)
    & 20.46 & 42.42 & 7.76 & 9.50 & 6.65 & 11.42 & 4.00 & 4.00 & 2.98 & 3.24 & 2.60 & 2.98 & 2.01 & 2.53 & 8.37 & 22.71 \\
    \hline
    RS-OVC (Ours)
    & 19.95 & 42.20 & \textbf{2.50} & \textbf{2.94} & 6.23 & 11.10 & 7.00 & 7.00 & 2.86 & 3.08 & 3.05 & 3.37 & 2.69 & 3.21 & 9.15 & 24.62 \\
    \hline
    \end{tabular}%
}

\label{tab:dior_results}
\end{table*}

Finally, \Cref{tab:rsocb_results} reports performance over the RSOC dataset\cite{gao2020counting}. RSOC includes buildings, vehicles and ships. Since ships and vehicles are included in our training set (\Cref{fig:data}), we perform evaluation only on the building class, that includes 1263 dense, low-resolution scenes (mean instance count of $26$ with s.t.d of $15.7$, see \cite{gao2020counting} for full statistics). In correlation with previous results, \Cref{tab:rsocb_results} shows that in these challenging scenes our method outperforms all considered baselines. While the OVD solution (LAE) excels in the simpler scenes of DIOR (\Cref{tab:dior_results}), it consistently struggles in complex multi-instance scenarios. The effect of scene-density over performance is further-analyzed in the supplementary.
\begin{table*}[t]
\centering
\setlength{\tabcolsep}{4pt}
\renewcommand{\arraystretch}{1.2}
\caption{Object-Counting performance on the RSOC dataset building class.}
\resizebox{\textwidth}{!}{%
    \begin{tabular}{|cc|cc|cc|cc|cc|}
    \hline
    \multicolumn{2}{|c|}{\textbf{CountGD}} & 
    \multicolumn{2}{c|}{\textbf{CountGD (RS-FT)}} & 
    \multicolumn{2}{c|}{\textbf{LAE}} & 
    \multicolumn{2}{c|}{\textbf{RS-OVC (RS-only)}} & 
    \multicolumn{2}{c|}{\textbf{RS-OVC (Ours)}} \\
    \hline
    \multicolumn{1}{|c}{MAE} & \multicolumn{1}{c|}{RMSE} & 
    \multicolumn{1}{c}{MAE} & \multicolumn{1}{c|}{RMSE} & 
    \multicolumn{1}{c}{MAE} & \multicolumn{1}{c|}{RMSE} & 
    \multicolumn{1}{c}{MAE} & \multicolumn{1}{c|}{RMSE} & 
    \multicolumn{1}{c}{MAE} & \multicolumn{1}{c|}{RMSE} \\
    \hline
    13.52 & 17.80 & 12.38 & 18.39 & 15.39 & 21.84 & 10.94 & 15.91 & \textbf{10.51} & \textbf{15.30} \\
    \hline
    \end{tabular}%
}

\label{tab:rsocb_results}
\end{table*}

\subsection{Qualitative Results}
\label{qual}
In this section we evaluate our method qualitatively, visually-assessing RS-OVC's performance in three increasingly-complex counting scenarios, that extend beyond the basic few-shot counting task that is visualized in \Cref{fig:teaser}.
\subsubsection{Local Semantic Understanding}\label{sec:local} 
\begin{figure}
    \centering
    \includegraphics[width=0.8\linewidth]{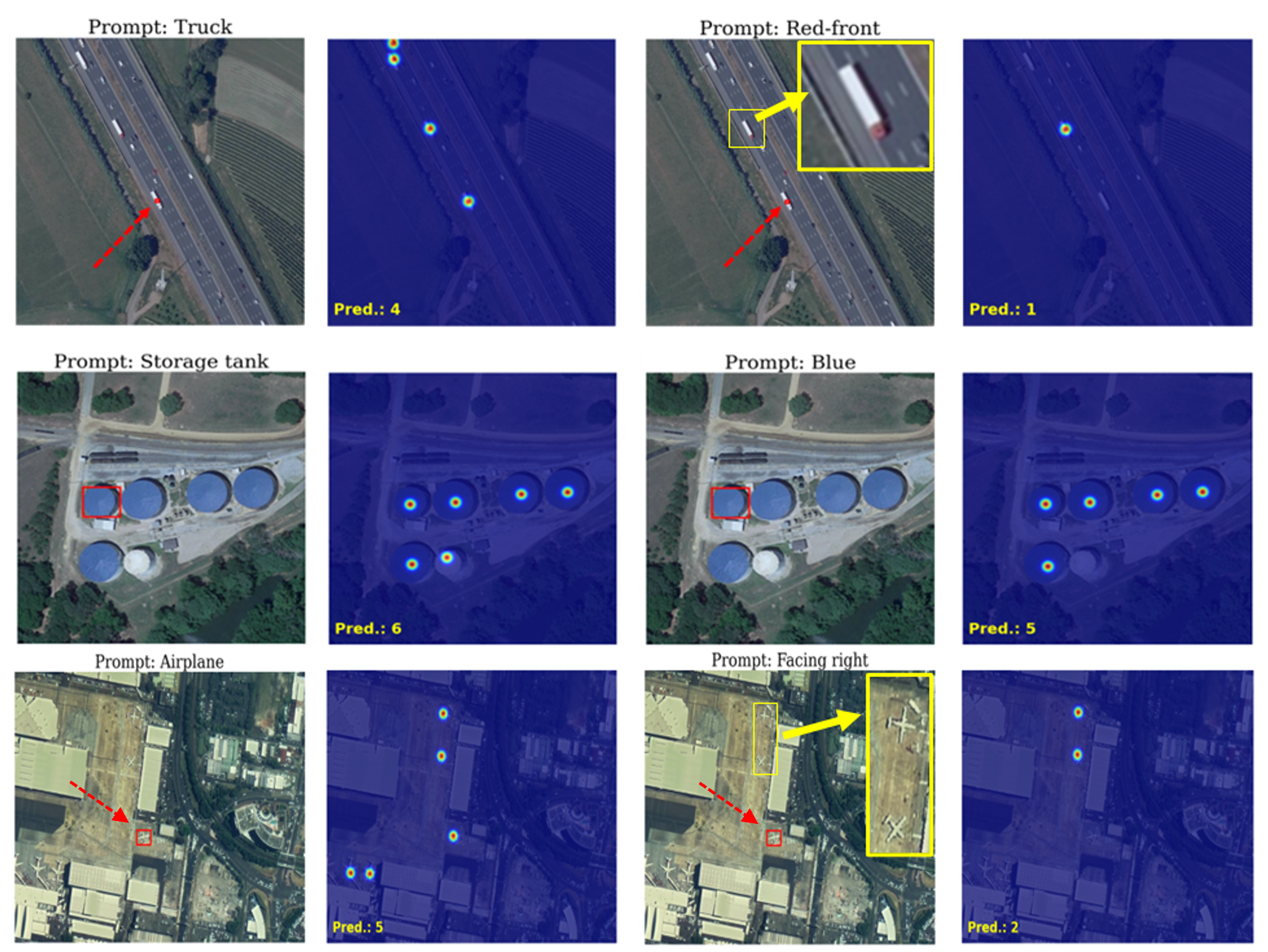}
    \caption{\textbf{Object confidence maps} - for the standard setting and for joint visual–textual conditioning. Correct predictions require local object-level semantic understanding. For instance, in the top row our model correctly selects the single truck that has a red front (and not other trucks or red cars) given the textual prompt. Red markings indicate visual exemplars. Yellow markings upscale important image regions. }
    \label{fig:semantic}
\end{figure}

- 
We evaluate our model's local semantic understanding, at the single-object level. Namely, how well RS-OVC grasps concepts that apply to a single object, rather than to a global environment. \Cref{fig:semantic} exhibits 3 rows representing 3 testing scenarios. In each row, the two leftmost images showcase the basic counting scenario of our method (similarly to \Cref{fig:teaser}) - for each image, the model is given a single visual exemplar and a matching text prompt, and it must count all objects matching these two correlated prompts. \\
The core of the experiment in this section is the two rightmost columns, where we feed our model with two different prompts - one given as a visual exemplar (e.g. a truck in the first row) and one as text (e.g. "red-front" in the same experiment). Our model must employ semantic object understanding in order to find the objects most-correlated with both prompts. Red boxes indicate visual exemplars (accompanied by red arrow pointers for small exemplars).\\
In the first row, our model correctly identifies the one truck with a red front (rightmost column), despite the features corresponding to the color "red" occupy little area in the image. Notably, objects that match only one condition (e.g., non-red trucks or red vehicles of other categories) are suppressed (although such objects appear in the image). Similarly, in the second row the model selects only blue storage tanks. In the third row  it correctly associates an orientation-specific textual prompt with the appropriate airplane instances.

\subsubsection{Global Semantic Understanding}\label{sec:global} 
\begin{figure}
    \centering
    \includegraphics[width=0.8\linewidth]{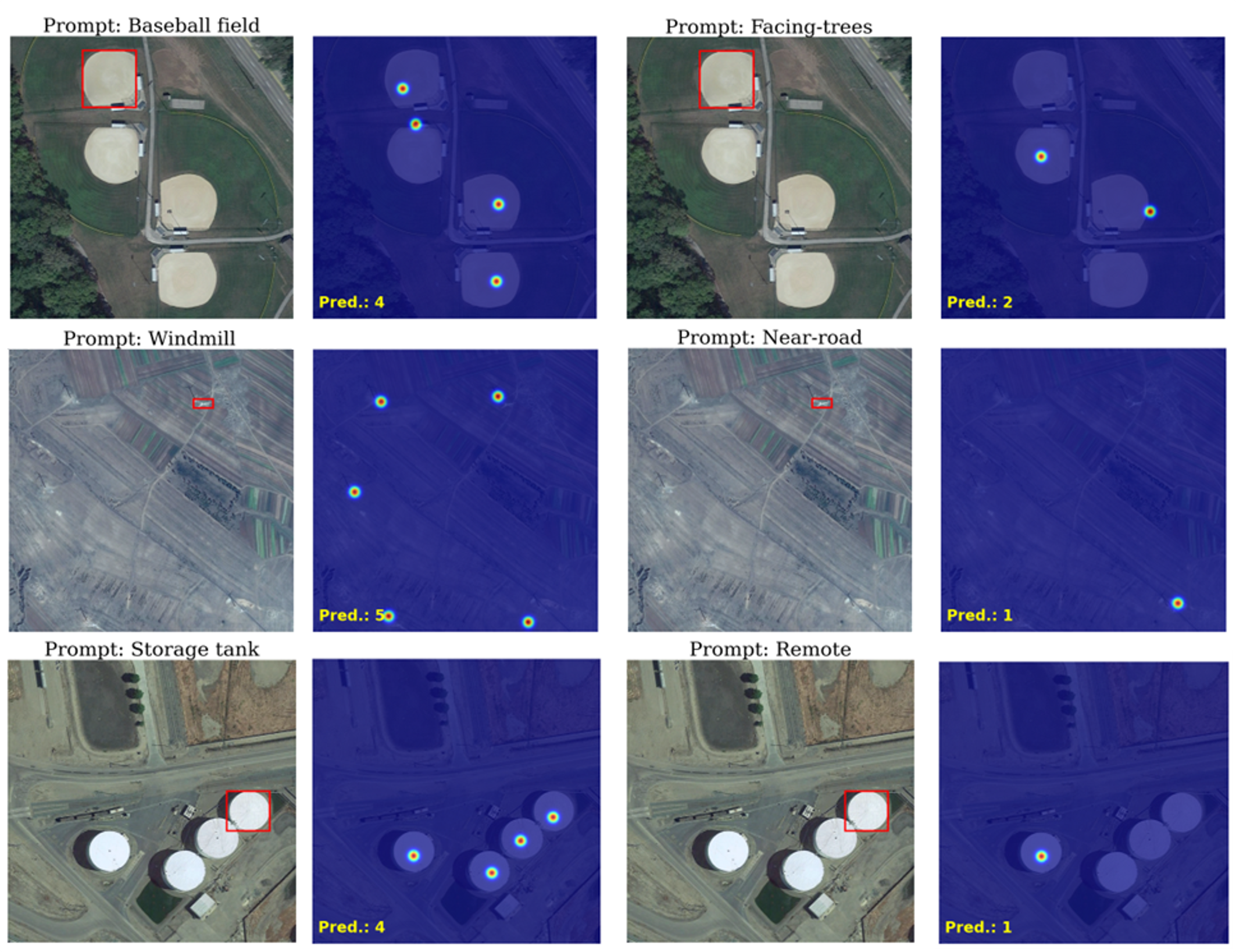}
     \caption{\textbf{Object confidence maps} - for the standard setting and for joint visual–textual conditioning. Correct predictions require global scene-level semantic understanding. The two rightmost columns introduce textual prompts that require relational reasoning. For instance, in the top row the model must identify absolute and relative spatial positions and orientations of baseball fields and trees to count correctly.}
    \label{fig:relations}
\end{figure}

-
We next assess our model's \textit{global} semantic understanding, focusing on its ability to perform global feature attribution and reason over relationships between multiple object instances across the scene. Results are shown in \Cref{fig:relations} - Unlike the previous section, focused on local understanding, in this section the text prompts are designed so that the correct output cannot be evaluated from looking at every object in isolation.\\
In the top row, the combination of the visual prompt \textit{baseball-field} with the textual prompt \textit{"facing trees"} requires the model to identify baseball-fields and trees, and determine the spatial relationship of each group of trees to each field. The model successfully selects only the baseball fields that satisfy this relational constraint. Similar behavior is observed in the second row, where the model identifies windmills located near roads.
The third row demonstrates a more complex scenario, in which the model must infer relative proximity among multiple instances to identify a remote container. Here, correct identification requires reasoning over inter-object distances rather than absolute spatial cues. Together, these examples indicate that RS-OVC acquires non-trivial relational understanding that extends beyond independent object recognition.

\subsubsection{Basic Reasoning} 
\begin{figure}
    \centering
    \includegraphics[width=0.8\linewidth]{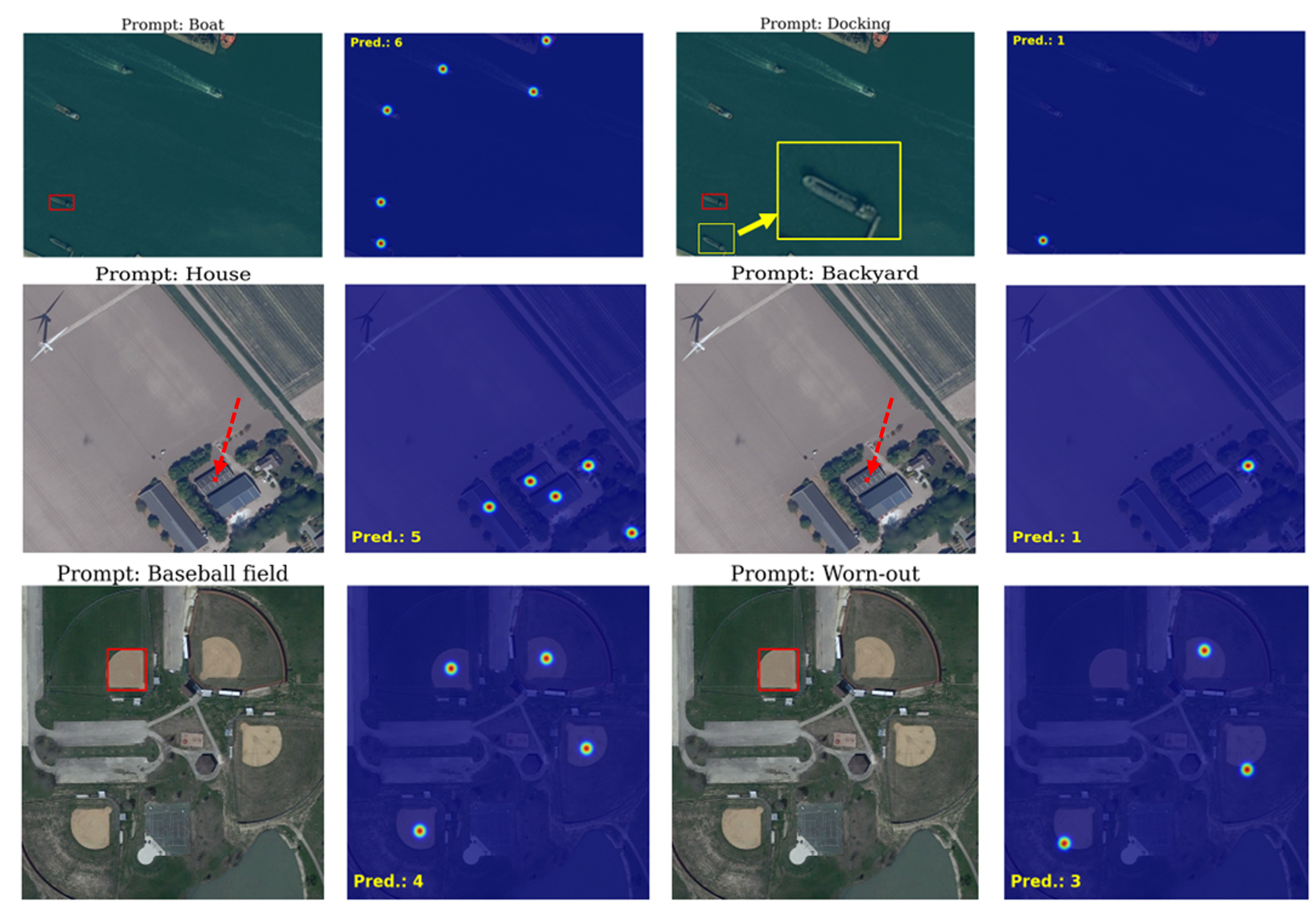}
    \caption{\textbf{Object confidence maps} -  for the standard setting and for joint visual–textual conditioning that require basic reasoning. For instance, in the first row the model must attribute the bottom boat's proximity to a pier to infer it is docking.}
    \label{fig:reasoning}
\end{figure}

-
Finally we show visual evidence (\Cref{fig:reasoning}) for our model's ability to leverage demonstrated scene-understanding capabilities to infer information that is not directly encoded within the visual features found in the scene - i.e. incorporate some level of reasoning into the counting task. In the first two columns of \Cref{fig:reasoning} we again present performance in the basic counting task (where visual and textual prompts are aligned).In the rightmost columns we evaluate reasoning-based prompts. In the first row, the model is given a visual prompt (\textit{"boat"}) and the textual prompt (\textit{"docking"}). Correct identification requires: (i) detecting boats, (ii) identifying contextual structures relevant to the text-prompt (e.g., piers), (iii) understanding spatial relationships between them, and (iv) inferring that proximity to a pier (and absence of motion cues) implies docking. The model successfully isolates the correct instance. Importantly, this knowledge required for reasoning is most-likely originated from the rich semantic understanding encapsulated within the latent space of the text encoder used by our model. The knowledge acquired during training mostly affects the ability to utilize this knowledge and correctly fuse it with the scene and the given conditioning supplied by the user.\\
The second row presents a similar reasoning task, where identifying a backyard requires associating grass regions with nearby houses. In the third row, the description of \textit{"worn-out"} must be reasoned from context - the three fields retrieved are worn-out when compared to the relatively greener field on the top-left.

\section{Conclusions \& Future Work}
\label{future_work}
In this work, we introduced RS-OVC - the first open-vocabulary counting framework for the domain of remote-sensing. Our approach leverages cross-domain feature fusion and counting backbones trained on rich and diverse data from general computer-vision, combined with a curated multi-dataset training regime in order to address the challenge of limited scale and semantic diversity of existing RS counting datasets, which hinders open-world counting in aerial imagery. Our evaluation places RS-OVC as superior to a representative OVD solution and to naive counting adaptations to the RS domain in complex, dense scenes, whilst showing a slight advantage for OVD in sparse and simpler scenes.\\

In spite of promising initial results, several limitations remain. First, our method exhibits on-par or inferior performance to the tested OVD baseline model in counting objects that appear in larger scales or lower density. We believe that our model's performance is, to some extent, hindered by our choice to favor simplicity and resource efficiency in RS finetuning. We are hopeful that future advancements in available datasets and methods for RS detection and counting would help push OVC in RS further. Furthermore, our method inherits the limitation of the CountGD framework we rely on, where the final predicted count is determined by a static threshold that has to be tuned using a designated validation set. We leave the exploration of automated solutions for future endeavors.

\bibliographystyle{splncs04}
\bibliography{bib}

\clearpage

\appendix
\section{Experimental Setup \& Implementation Details}
\subsection{RS-OVC Implementation}
Our implementation of RS-OVC mostly follows the official code published in the CountGD paper, which is the backbone for our model. Our feature-injection module (Section 3.2) is implemented using the standard Pytorch \textit{nn.Module} implementation with $8$ attention heads. Every hierarchy level of Swin-transformer features $z_{CV}$ undergoes bilinear interpolation to the spatial dimensions of the RS encoder's latent space ($z_{RS}$). Channel dimensions of $z_{CV}$ are project to those of $z_{RS}$ using 2D-convolutional projection with $1\times1$ kernels. \\
In our curation of the FAIR-1M dataset, we separate the \textit{ship} class to two semantically distinct object classes - \textit{ships} (used for training) include FAIR-1M's \textit{Warship, Cargo-ship, Engineering-ship and Passenger-ship} native classes. The \textit{boat} class (used for evaluation) includes FAIR-1M's \textit{Tugboat, Motorboat and Fishing-boat} object classes.
\subsection{Baseline Experimental Setup}

The CountGD and LAE baselines are used off-the-shelf (up to adaptation of the detection thresholds, similarly to RS-OVC). We use the same textual prompts as the ones used in RS-OVC for all evaluations. For LAE we use the final number of detected bounding boxed as the final count prediction. We evaluate CountGD (and all other few-shot baselines) with the same visual exemplars as the ones fed into RS-OVC. 

\section{Optimization Details}
CountGD and LAE are used off-the-shelf without any finetuning. Each of RS-OVC, CountGD-RSFT and the RS-only baselines is optimized over our curated training data (Section 3.3) for 30 epochs using the Adam optimizer with a learn-rate of $1e-5$ and batch-size of $4$. All experiments in this paper were conducted on four NVIDIA-A100 GPUs for training and a single NVIDIA-A100 GPU for evaluation. \\
All counting models are initialized with the original checkpoints from the CountGD paper. In all experiments, all image and text encoders remain frozen (their parameters are not optimized). For RS-OVC and finetuned CountGD variants we optimize only the decoder and feature-fusion modules. \\
The detection thresholds for evaluation are determined by randomly sampling $100$ images from the curated training set. These validation samples are excluded from the final training set. Furthermore, in this work we found it beneficial to use adaptive thresholding in our zero-shot experiments (reported later in the appendix). To do so we triple the detection threshold for any similarity matrix where the maximum value (after Sigmoid) is larger than some constant predefined $\tau$. The value of $\tau$ is determined by the same validation set, and is applicable only in zero-shot inference.

\section{Additional Results}
\subsection{Zero-shot Performance}
The main paper mainly presents our method within 3-shot or single-shot scenarios (i.e. one or three visual exemplars). In this section we explore RS-OVC's performance in a zero-shot setup. Namely, we repeat all experiments without visual-exemplars fed into our model. Results are brought in \Cref{tab:nwpu_zs,tab:fair_zs,tab:dota_zs,tab:dior_zs,tab:rsocb_zs}. Each table reports RS-OVC's zero-shot performance alongside performance of the full (3-shot) RS-OVC and our strongest baseline (LAE), reported in the main paper. \\
In terms of comparison with the OVD baseline (LAE), results are correlated with those from the main paper - For the NWPU-MOC, FAIR-1M and RSOC-Building datasets, that are mostly characterized by dense scenes and low-resolution objects, our method's zero-shot performance outperforms that of LAE. Namely, RS-OVC's advantage over LAE is not derived purely from its usage of visual exemplars in a few-shot setting. In datasets where objects are generally larger and scenes less dense, like DOTA and DIOR, the performance gap is more nuanced. While LAE achieves a higher overall mean performance, the advantage shifts between zero-shot RS-OVC and LAE depending on the specific object class.\\
In terms of comparison between zero-shot and three-shot performance, \Cref{tab:nwpu_zs,tab:fair_zs,tab:dota_zs,tab:rsocb_zs} demonstrate of performance boost of $25-50\%$ when transitioning from zero-shot to three-shot RS-OVC. These results emphasize the advantage of our method, that supports few shot inference, over strictly zero-shot baselines. 

\begin{table}[t]
\centering
\setlength{\tabcolsep}{2pt} 
\renewcommand{\arraystretch}{1.2}
\caption{Zero-shot object-Counting performance on NWPU-MOC test classes, compared to 3-shot RS-OVC and LAE. Our method outperforms LAE in the zero-shot setup.}
\resizebox{\linewidth}{!}{%
    
    \begin{tabular}{|l|cc|cc|cc|cc|cc|cc|} 
    \hline
    \textbf{Method}
    & \multicolumn{2}{c|}{Boat}
    & \multicolumn{2}{c|}{House}
    & \multicolumn{2}{c|}{Truck}
    & \multicolumn{2}{c|}{Stadium}
    & \multicolumn{2}{c|}{Airplane}
    & \multicolumn{2}{c|}{Mean} \\
    \cline{2-13}
    & MAE & RMSE & MAE & RMSE & MAE & RMSE & MAE & RMSE & MAE & RMSE & MAE & RMSE \\
    \hline
    LAE 
    & 40.03 & 77.90 & 27.11 & 38.73 & 12.72 & 20.70 & 5.50 & 5.83 & 4.96 & 5.21 & 23.44 & 39.73 \\
    \hline
    RS-OVC - zero-shot (Ours)
    & 32.04 & 68.23 & 21.98 & 34.72 & 7.43 & 16.43 & 18.66 & 24.99 & 1.31 & 2.06 & 18.46 & 35.22 \\
    \hline
    RS-OVC (Ours)
    & 7.5 & 15.44 & 15.52 & 26.87 & 8.74 & 24.2 & 5.52 & 6.37 & 3.08 & 3.55 & 12.42 & 24.72\\
    \hline
    \end{tabular}%
}

\label{tab:nwpu_zs}
\end{table}

\begin{table}[t]
\centering
\footnotesize
\setlength{\tabcolsep}{4pt}
\renewcommand{\arraystretch}{1.2}
\caption{Zero-shot performance on FAIR-1M test classes.}
\resizebox{\linewidth}{!}{

\begin{tabular}{|l|cc|cc|cc|cc|cc|cc|cc|}
\toprule
\textbf{Method}
& \multicolumn{2}{c|}{Tractor}
& \multicolumn{2}{c|}{Boat}
& \multicolumn{2}{c|}{Truck}
& \multicolumn{2}{c|}{Airplane}
& \multicolumn{2}{c|}{Baseball-Field}
& \multicolumn{2}{c|}{Mean} \\
\cline{2-13}
& MAE & RMSE & MAE & RMSE & MAE & RMSE & MAE & RMSE & MAE & RMSE & MAE & RMSE  \\
\midrule
LAE
& 41.58 & 89.57 & 19.16 & 45.53 & 10.65 & 15.88 & 1.29 & 6.74 & 0.48 & 0.82 & 7.56 & 19.68 \\
\hline
RS-OVC-zero-shot (Ours)
& 36.84 & 85.36 & 18.40 & 41.97 & 6.01 & 11.88 & 6.08 & 11.07 & 5.90 & 7.49 & 7.51 & 18.31 \\
\hline
RS-OVC (Ours)
& 26.84 & 47.98 & 12.79 & 27.45 & 5.54 & 10.41 & 4.21 & 10.22 & 3.00 & 3.06 & 5.81 & 13.58 \\
\bottomrule
\end{tabular}
}

\label{tab:fair_zs}
\end{table}

\begin{table*}[t]
\centering
\setlength{\tabcolsep}{2pt} 
\renewcommand{\arraystretch}{1.2}
\caption{Zero-shot performance on DOTA test classes.}
\resizebox{\textwidth}{!}{%
    \begin{tabular}{|l|cc|cc|cc|cc|cc|cc|cc|} 
    \hline
    \textbf{Method}
    & \multicolumn{2}{c|}{Storage Tank}
    & \multicolumn{2}{c|}{Airplane}
    & \multicolumn{2}{c|}{Bridge}
    & \multicolumn{2}{c|}{Soccer Ball Field}
    & \multicolumn{2}{c|}{Roundabout}
    & \multicolumn{2}{c|}{Baseball Diamond}
    & \multicolumn{2}{c|}{Mean} \\
    \cline{2-15}
    & MAE & RMSE & MAE & RMSE & MAE & RMSE & MAE & RMSE & MAE & RMSE & MAE & RMSE & MAE & RMSE \\
    \hline
    LAE
    & 57.18 & 121.67 & 18.86 & 59.40 & 14.80 & 23.91 & 7.21 & 9.01 & 6.82 & 8.06 & 4.67 & 6.69 & 23.59 & 67.94 \\
    \hline
    RS-OVC-zero-shot (Ours)
    & 51.62 & 113.30 & 35.81 & 77.73 & 12.64 & 21.04 & 14.54 & 26.44 & 8.16 & 21.37 & 5.62 & 9.63 & 27.69 & 67.79 \\
    \hline
    RS-OVC (Ours)
    & 49.76 & 111.28 & 24.13 & 41.66 & 11.51 & 18.61 & 5.50 & 6.27 & 5.67 & 6.93 & 5.33 & 6.65 & 23.16 & 58.15 \\ 
    \hline
    \end{tabular}%
}
\label{tab:dota_zs}
\end{table*}

\begin{table*}[t]
\centering
\setlength{\tabcolsep}{2pt} 
\renewcommand{\arraystretch}{1.2}
\caption{Zero-shot performance on DIOR test classes.}
\resizebox{\textwidth}{!}{%
    \begin{tabular}{|l|cc|cc|cc|cc|cc|cc|cc|cc|} 
    \hline
    \textbf{Method}
    & \multicolumn{2}{c|}{Storage Tank}
    & \multicolumn{2}{c|}{Bridge}
    & \multicolumn{2}{c|}{Airplane}
    & \multicolumn{2}{c|}{Stadium}
    & \multicolumn{2}{c|}{Chimney}
    & \multicolumn{2}{c|}{Baseball Field}
    & \multicolumn{2}{c|}{Windmill}
    & \multicolumn{2}{c|}{Mean} \\
    \cline{2-17}
    & MAE & RMSE & MAE & RMSE & MAE & RMSE & MAE & RMSE & MAE & RMSE & MAE & RMSE & MAE & RMSE & MAE & RMSE \\
    \hline
    LAE
    & 12.80 & 36.24 & 4.15 & 5.00 & 1.16 & 3.24 & 3.00 & 3.00 & 1.09 & 2.17 & 1.14 & 2.79 & 0.89 & 2.12 & 5.05 & 20.68 \\
    \hline
     RS-OVC-zero-shot (Ours)
    & 21.29 & 42.98 & 2.59 & 4.00 & 7.25 & 12.48 & 0.00 & 0.00 & 1.48 & 2.05 & 1.85 & 2.97 & 2.15 & 3.20 & 9.50 & 25.20 \\
    \hline
    RS-OVC (Ours)
    & 19.95 & 42.20 & 2.50 & 2.94 & 6.23 & 11.10 & 7.00 & 7.00 & 2.86 & 3.08 & 3.05 & 3.37 & 2.69 & 3.21 & 9.15 & 24.62 \\
    \hline
    \end{tabular}%
}

\label{tab:dior_zs}
\end{table*}

\begin{table*}[t]
\centering
\setlength{\tabcolsep}{4pt}
\renewcommand{\arraystretch}{1.2}
\caption{Zero-shot performance on the RSOC dataset building class.}
\resizebox{0.6\textwidth}{!}{%
    \begin{tabular}{|cc|cc|cc|}
    \hline
    \multicolumn{2}{|c|}{\textbf{LAE}} & 
    \multicolumn{2}{c|}{\textbf{RS-OVC-zero-shot (Ours)}} & 
    \multicolumn{2}{c|}{\textbf{RS-OVC (Ours)}} \\
    \hline
    \multicolumn{1}{|c}{MAE} & \multicolumn{1}{c|}{RMSE} & 
    \multicolumn{1}{c}{MAE} & \multicolumn{1}{c|}{RMSE} & 
    \multicolumn{1}{c}{MAE} & \multicolumn{1}{c|}{RMSE} \\
    \hline
    15.39 & 21.84 & 15.16 & 21.49 & 10.51 & 15.30 \\
    \hline
    \end{tabular}%
}
\label{tab:rsocb_zs}
\end{table*}

\subsection{Analysis of Scene Density Effect}
\begin{figure}
    \centering
    \includegraphics[width=\linewidth]{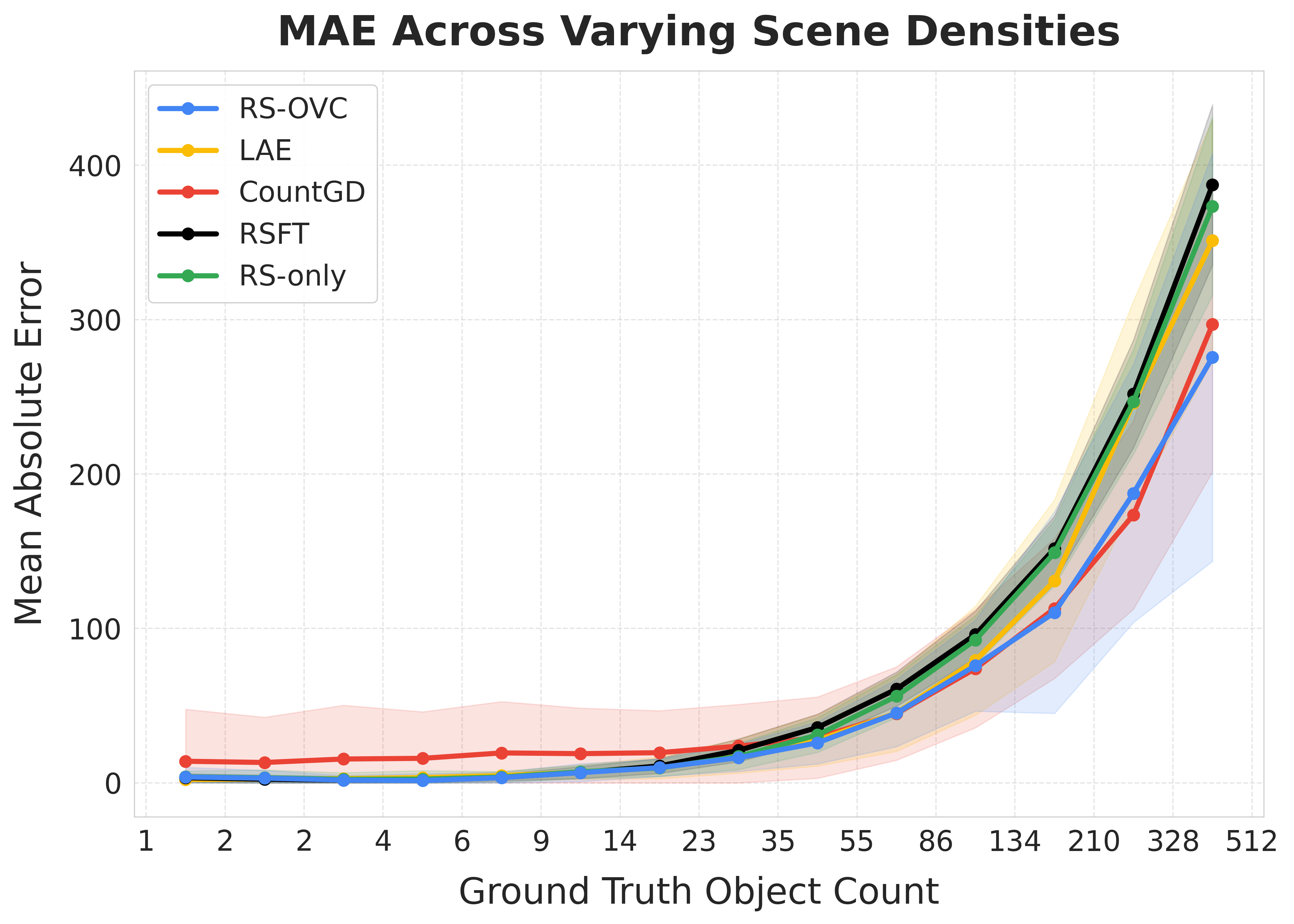}
    \caption{\textbf{MAE as a function of number of instances in the scene} - aggregated across NWPU-MOC, FAIR-1M, DOTA, DIOR and RSOC-BUilding datasets. Data points are aggregated into logarithmically spaced bins to visualize performance across widely varying object counts, with shaded regions indicating standard deviation. We report results for RS-OVC (our method, in blue), LAE (OVD baseline, in yellow), off-the-shelf CountGD (without any RS adaptation, in red), CountGD finetuned over RS data (RSFT, black) and CountGD finetuned over RS data with a pre-trained RS-designated image encoder (RS-only, green).}
    \label{fig:density}
\end{figure}

One key claim made in the main paper is that our method's advantage over OVD is most substantial in dense scenes with high instance counts. To illustrate this, in \Cref{fig:density} we plot the aggregated MAE across all test classes from the NWPU-MOC, FAIR-1M, DOTA, DIOR and RSOC-Building datasets, as a function of ground-truth object count in the scene. Due to the high variance of this function, we bin all ground-truth count values to 15 bins such that every bin has the same number of samples attributed to it (i.e. quantile binning).\\
\Cref{fig:density} clearly illustrates our previously-mentioned claim. For sparser scenes (left side of the plot), our method is on-par with LAE and other baselines. As the scene density grows, we observe clear divergence between the OVD baselines (yellow) and RS-OVC (in blue). Results also further-illustrate the importance of the finetuning method and feature-injection used i RS-OVC - the two "naive" adaptations of CountGD to the RS domain (green and black) are on-par with RS-OVC and LAE for sparser scenes, however their performance drastically deteriorates as scenes become denser (similarly to LAE). Conversly, the off-the-shelf CountGD baseline (in red) is on-par with RS-OVC for dense scenes, however it is far-inferior to all other baselines in sparser settings. Our method is the only one that manages to maintain its best performance across all density levels.

\end{document}